# Digital-twin-enhanced metal tube bending forming real-time prediction method based on Multi-source-input MTL

Chang Sun[1], Zili Wang[1,2*], Shuyou Zhang[1,2], Taotao Zhou[1], Jie Li[1], Jianrong Tan[1,2]

*(1. State Key Laboratory of Fluid Power and Mechatronic Systems, Zhejiang University, Hangzhou, 310027, China. 2. Engineering Research Center for Design Engineering and Digital Twin of Zhejiang Province, Zhejiang University, Hangzhou, 310027, China)*

*Corresponding author:* Zili Wang, e-mail: ziliwang@zju.edu.cn.

**Abstract**

As one of the most widely used metal tube bending methods, the rotary draw bending (RDB) process enables reliable and high-precision metal tube bending forming (MTBF). The forming accuracy is seriously affected by the springback and other potential forming defects, of which the mechanism analysis is difficult to deal with. At the same time, the existing methods are mainly conducted in offline space, ignoring the real-time information in the physical world, which is unreliable and inefficient. To address this issue, a digital-twin-enhanced (DT-enhanced) metal tube bending forming real-time prediction method based on multi-source-input multi-task learning (MTL) is proposed. The new method can achieve comprehensive MTBF real-time prediction. By sharing the common feature of the multi-close domain and adopting group regularization strategy on feature sharing and accepting layers, the accuracy and efficiency of the multi-source-input MTL can be guaranteed. Enhanced by DT, the physical real-time deformation data is aligned in the image dimension by an improved Grammy Angle Field (GAF) conversion, realizing the reflection of the actual processing. Different from the traditional offline prediction methods, the new method integrates the virtual and physical data to achieve a more efficient and accurate real-time prediction result. and the DT mapping connection between virtual and physical systems can be achieved. To exclude the effects of equipment errors, the effectiveness of the proposed method is verified on the physical experiment-verified FE simulation scenarios. At the same time, the common pre-training networks are compared with the proposed method. The results show that the proposed DT-enhanced prediction method is more accurate and efficient.

Keywords: Metal tube bending forming (MTBF); Digital twin (DT); Multi-source-input MTL; L1/L2 regularization; GAF

## 1 Introduction

Metal bent-tube plays an irreplaceable role in the industry. Rotate draw bending (RDB) processing is one of the most reliable and stable approaches to conduct metal tube bending forming (MTBF), which has the advantages of high precision, low cost, and flexible processing. However, with the removal of the molds constraints after processing, the elastic deformation will recover under the effect of internal processing stress, resulting in the so-called "springback" phenomenon. Springback will significantly affect the bent-tube axis accuracy, and further affect the assembly performance. One of the state-of-the-art approaches dealing with springback is to set a reversed bending compensation, guaranteeing the product final shape after springback, which makes the need for springback prediction necessary. At the same time, there are still other defects that might occur during processing, such as the cross-section distortion, and the bending wrinkling. These potential defects can also affect the bending process and determines the final shape together with springback, which needs to be inhibited.

From the perspective of stress-strain state analysis (Zhan et al., 2006), bending processing modes (Ahn et al., 2020; Zhai et al., 2018) and material models (Zhan et al., 2016), the mechanism of springback has been preliminarily raveled. However, deformation coupling is tremendously complex, making the existing methods difficult to be applied in engineering applications.

With the help of finite element (FE) simulation (Baseri et al., 2011; Jamli et al., 2014) and machine learning (Zhou et al., 2021), the effect of the shape and process parameter have been further studied. At the same time, similar methods have also been applied to the defects (Li et al., 2014). However, the traditional methods cannot achieve the regression of the springback and classification of the defects at the same time. With the high cost of the defect dataset acquisition, enough MTBF information cannot be always guaranteed. The springback and potential defects in MTBF multi-close domain tasks. Their coupling relationship is extremely complex, and is all essentially caused by tube deformation. MTL (Caruana, 1997) can share the common features between them, realize the multi-task prediction, and significantly reduce the cost (Yang et al., 2013). However, based on the known-parameters before processing, the above-all methods are mainly implemented in offline space, which ignores the real-time information in the real world.

Digital twin (DT) technology is proposed in recent decades to deeply connect the virtual system and physical system, effectively promoting the development of smart factories and manufacturing (Wang and Luo, 2021). DT improves the efficiency of the full life cycle such as machining (Xia et al., 2021), running status monitoring (Hu et al., 2021), and fault diagnosis (Tao et al., 2018b). Especially, DT can simulate and monitor processing in real-time by establishing a connection between the virtual and physical system (Soderberg et al., 2017), providing a way to integrate the virtual and physical data in MTBF prediction. However, to construct the real-time connections, DT often requires large amounts of data acquired by sensors (Uhlemann et al., 2017). This can limit the field of DT vision, making the existing methods easy to ignore important data features, and difficult to consider the uncertainty factors in the actual process. Since the implementation of the existing DT is mainly based on machine learning, the physical real-time information needs to be pre-processed and limited by the data dimension. Methods to pre-process series data commonly used in machine learning, such as padding (Yatbaz et al., 2021) and series elongation (Min et al., 2019), limit the real-time information in a fixed period, which have the risk of data distortion and information loss, and affect the prediction accuracy.

To alleviate the above-mentioned difficulties, a DT-enhanced metal tube bending forming real-time prediction method based on multi-source-input multi-task learning (MTL) is proposed. In fact, once the process plan of MTBF is known, a preliminary prediction result can be obtained, which is the existing common method. The process plan is essentially belonging to virtual data, which can be generated, stored, and utilized in the offline space. Correspondingly, the real-time deformation data in the MTBF process can be measured by sensors, reflecting the physical world in real-time. The virtual data and physical data will be integrated by the proposed method. Through the Gramian Angular Fields (GAF) conversion, the real-time physical deformation data will be used together with virtual data, especially with shape and design parameters, as multiple sources for the comprehensive prediction of the bent-tube final shape. Multi-source-input MTL adopts the form of multi-input and multi-output with soft parameter sharing architecture, combine with L1/L2 group regularization strategy to make the model more accurate and robust.

Different from the common offline space methods, the proposed method can integrate the virtual and physical data, to achieve more efficient, accurate and real-time prediction based on the DT mapping connection between virtual and physical systems. The Multi-source-input MTL can comprehensively predict the bent-tube, which achieves the regression of the springback and classification of the defects at the same time. By sharing the common feature of the multi-close domain, the performance and efficiency of the network are improved. Enhanced by DT, the proposed method can integrate the different information sources from the physical and virtual data, to realize the feedback of the real-time processing. The effectiveness of the proposed methods is carried out in a 60601-Al tube bending case with physical experiment-verified FE simulation scenarios. The common pre-training networks are also compared with our proposed network. The results show that the proposed DT-enhanced multi-source-input MTL prediction method is more accurate and efficient.

The rest of this paper is organized as follows. Section 2 introduces the related works. In Section 3, the DT-enhanced RDB metal tube bending axis accuracy real-time prediction framework is put forward. The multi-source-input MTL prediction model is constructed in Section 4, which integrates the GAF conversion pre-processing and L1/L2 group regularization strategy with the multi-source-input MTL model. To verify the feasibility of the proposed method, a 60601-Al tube bending case study is carried out in physical experiment-verified FE simulation scenarios in Section 5. Finally, the conclusions are drawn in Section 6. The codes of this work are available in https://github.com/sooncheer0420/SaMo_MTL.git.

## 2 Related works

### 2.1 The traditional product accuracy analysis method of RDB

RDB process is a stretch bending forming method, which will end up with obvious elastoplastic deformation in tubes. With the removal of the load applied by molds after processing, the elastic deformation and part of the plastic deformation will recover, resulting in further deformation in the axial direction, which is called the "springback". With the hollow structure, the cross-section of the blank tube will be affected by the material flow and the radial stress during bending processing, resulting in the potential risk of collapse and wrinkling. The springback and potential defects can have a huge impact on the axial accuracy and the quality of the product.

With the stress-strain state analysis, the mechanism of springback can be raveled. Zhan et al. (2006) preliminarily analyzed the bent-tube deformation mechanism. On this basis, further studies have been made in tube blank pre-processing and bending models (Ahn et al., 2020; Zhai et al., 2018). The influence of Young's modulus, yield stress and strain hardening index on production have also been studied (Zhan et al., 2016), and a variety of springback analysis models have been derived (Jiang et al., 2010; Liu et al., 2013; Shahabi and Nayebi, 2015; Zhu et al., 2013). At the same time, Wang et al. (2022) considered the effect of cross-section distortion when studying springback of tube bending, and similar work was also shown in (Zhou et al., 2021). FE simulation is another effective method. The influence of the processing parameters can be qualitatively analyzed (Li et al., 2020). Machine learning, which passes parameters through nonlinear functions to reflect the inherent relationship of a large number of high-dimensional data, also performs well combined with the FE simulation (Baseri et al., 2011; Jamli et al., 2014).

However, the deformation coupling is tremendously complex, which still needed to be further studied. Besides, in the existing methods, the prediction-needed parameters are mainly determined before processing, which means the prediction results are mainly concluded from the processing plan. Based on the known-parameters before processing, the existing methods are mainly implemented in offline space. However, it ignores the real-time information in the physical system. Since the unreliable acquisition of certain parameters, such as friction and gap between molds and tube blank, the prediction can also be unreliable.

## 2.2 MTL for prediction tasks

Relying on sufficient datasets, traditional machine learning can achieve the springback or single defect prediction. However, the traditional methods cannot achieve the regression of the springback and classification of the defects at the same time. Besides, with the high cost of the defect dataset acquisition, enough MTBF information cannot be always guaranteed.

MTL (Caruana, 1997) can identify and share the common features between the multi-close domains, realize the multi-task prediction, and significantly reduce the cost (Yang et al., 2013). Considering multi-dataset-noise interference, parameters eavesdropping (Abu-Mostafa, 1990), and representation bias (Baxter, 2000), MTL can better prevent over-fitting, and has a better training effect as well as generalization performance (Li et al., 2016). Various configurations of MTL have been proposed (Lu et al., 2017; Misra et al., 2016; Sogaard and Goldberg, 2016). These special architectures of MTL enable it to measure a variety of input features at the same time. Bai et al. (2018) presented a multi-task generative adversarial network to handle the detection problem of small objects. Zhang et al. (2020) integrated LSTM into MTL to predict the short-term passenger demand at a multi-zone level.

With the same dataset, MTL has been empirically and theoretically proved to have a better performance than learning independently (Zhang and Yang, 2018). Zhang et al. (2014) used MTL for facial detection and achieved good results with head pose estimation and facial attribute inference. Similarly, the springback and potential defects in MTBF multi-close domain tasks. Their coupling relationship is extremely complex, and is all essentially caused by tube deformation. Through MTL, the deformation knowledge and the underlying characteristics of each task can be shared to improve the training efficiency and prediction accuracy.

However, based on the known-parameters before processing, the above-all methods are mainly implemented in offline space, which ignore the real-time information in the physical system. At the same time, based on the single virtual information source, traditional MTL cannot realize the integration and utilization of physical information and virtual data.

## 2.3 DT for real-time prediction

The concept of DT has constantly evolved in the recent decade. By connecting the physical and virtual worlds, DT technology enables the feature mapping of the individual processing to provide customized optimization and decision-making basis, which plays a vital role in the whole life cycle of the intelligent manufacturing. In the design phase, DT begins to require product-oriented data for the entire life cycle. Combining machine learning or optimization algorithms (Immonen, 2022), DTs have demonstrated effectiveness in product design (Schleich et al., 2017), production line design (Zhang et al., 2017a), and virtual commissioning (Fält and Halmsjö, 2016; Guerrero et al., 2014). In addition, DT has been applied in the processing stage such as the production equipment inspection management (Zhang et al., 2017b) and the production plan optimization (Min et al., 2019). Running state detection and state regulation could be one of the most widely used aspects of DTs (Hu et al., 2021; Qin et al., 2021; Schleich et al., 2017; Tao et al., 2018b).

To construct the real-time connection between the physical and virtual worlds, DT often requires large amounts of data acquired by sensors (Uhlemann et al., 2017). The transmission of large amounts of data requires huge computing resources, resulting in risks such as the discrepancy between virtual and physical manufacturing, or out-of-sync communications caused by hardware latency (Tao et al., 2018a). Although with the combination of machine learning (Xia et al., 2021) and the advanced computational model (Song et al. 2021), the requirement of the real-time measured data reduces, it has not been reduced to an acceptable range yet (Pan and Zhang, 2021). The cost of the feature quantity measurement in the processing can be expensive (Gunther et al., 2016). Meanwhile, the existing methods of using sensor data will limit the field of vision of DT to obtain information, making it difficult to consider the uncertainty factors in the actual processing.

Since the implementation of the existing DT is mainly based on machine learning, the physical real-time information limited by the data dimension needs to be pre-processed. However, in the common pre-processing methods, such as padding (Yatbaz et al., 2021) and series elongation (Min et al., 2019), the real-time information is limited to a fixed period, which have the risk of data distortion and information loss, resulting in some unexpected effects on the prediction. At the same time, DT-based machine learning architectures still needed to be implemented the capable of handling small datasets. The further development of DT in the field of product-oriented feature mapping is still limited.

## 3  DT-enhanced RDB real-time prediction framework of MTBF

The MTBF needs to be completed by the physical system, which includes the tube RDB machine tool and the sensor measurement device. It can be affected by multiple parameters and factors, as shown in Table 1. Specifically, the tube diameter $D$ and wall thickness $t$ of the tube blank is the basis of the product. The tube blank is forced to lean against the inner arc groove surface of the bending die. The radius $R_B$, rotation angle $\alpha_B$ and angular velocity $\omega_B$ of the bending die directly determine the final shape of the bent-tube. The initial position $L_P$ and the boost velocity $v_B$ of the pressure die can affect the preload and axial stress. Besides, the gap and friction between the molds and the tube blank also affect the quality significantly.

**Table 1** Main parameters in metal tube RDB

| Property | Parameter (Unit) | Abbreviation |
|---|---|---|
| Shape parameter | Tube diameter (mm) | $D$ |
|  | Wall thickness (mm) | $t$ |
| Design parameter | Radius of bending die (mm) | $R_B$ |
|  | Bending angle of bending die (rad) | $\alpha_B$ |
|  | Angular velocity of bending die (rad/s) | $\omega_B$ |
| Process parameter | The initial position of the pressure die (mm) | $L_P$ |
|  | Boost velocity of pressure die (mm/s) | $v_B$ |
|  | Friction between bending die and tube | $f_B$ |
|  | Friction between wiper die and tube | $f_W$ |
|  | Friction between pressure die and tube | $f_P$ |
|  | The gap between bending die and tube (mm) | $G_B$ |
|  | The gap between clamp die and tube (mm) | $G_c$ |
|  | The gap between pressure die and tube (mm) | $G_P$ |

As mentioned in Section 1, once the process plan of MTBF is known, a preliminary prediction result can be obtained, which is the existing common method. The process plan is essentially belonging to virtual data, which can be generated, stored and utilized in the offline space. The virtual data includes the shape parameter and the design parameter. It should be mentioned that even if the same molds are used according to the same process plan, the measurement of process parameters still can be unreliable due to the wear and the instability of the lubrication effect during processing. Since it is difficult and unreliable to obtain the process parameter, the prediction reliability can be affected. At the same time, the combined effect of these parameters and uncertain environmental factors will be directly reflected in the deformation of the bent-tube. Therefore, the real-time deformation data is taken as the physical data in the DT loop, which can be measured by sensors in the MTBF process, reflecting the effects of low-reliable parameters and uncertain factors of the physical world in real-time.

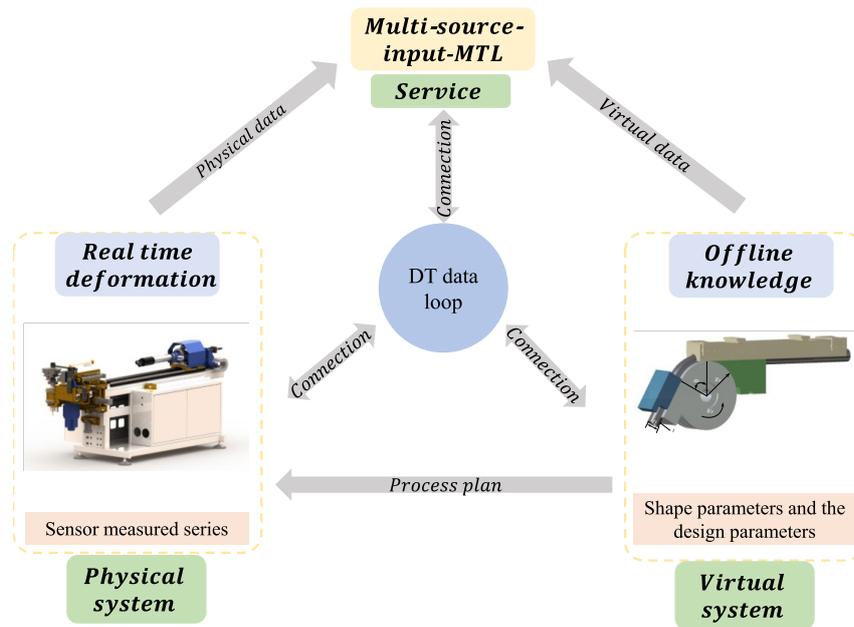

**Fig. 1** DT-enhanced tube RDB in five dimensions

Different from the common offline space methods, the proposed method can integrate the virtual and physical data to achieve more efficient and accurate real-time prediction based on the DT mapping connection between virtual and physical systems. Specifically, the DT-enhanced MTBF framework includes five dimensions (Tao et al., 2018a), as shown in Fig. 1. Virtual data serves as the basis for predictions, providing advanced end-to-end MTBF knowledge. The real-time data measured by the sensor is taken as the physical data to realize the reflection of the actual processing process. The process parameter will not be used directly, and its influence on the final product is reflected through the physical data, which is also an important measure to improve reliability. On this basis, the Multi-source-input MTL proposed in this paper integrates the physical and virtual data through the multi-source input architecture, achieving the comprehensive and accurate real-time prediction in the service dimension.

## 4  Multi-source-input MTL prediction method

### 4.1  The method summary

With the support of the DT data loop, the virtual data and physical data can be integrated to achieve real-time prediction of MTBF. Therefore, the Multi-source-input MTL prediction method is proposed, as shown in Fig. 2.

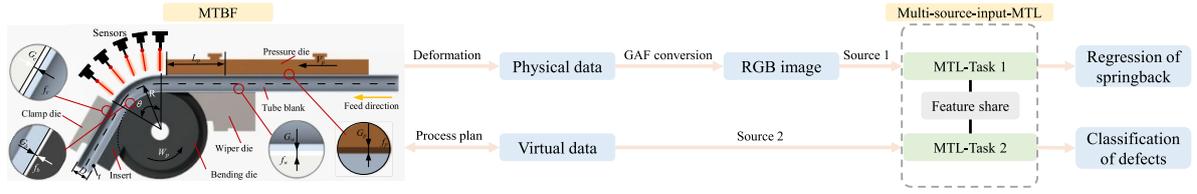

**Fig. 2** The data flow of the proposed Multi-source-input MTL prediction method

As mentioned in Section 3, the virtual data is clarified before processing for a set of MTBF. As processing begins, physical data will be measured to reflect real-time deformation. Including shape and design parameters, the virtual data dimensions remain the same. However, the physical data is continuously extended as the processing proceeds. Therefore, before being inputted into the deep learning framework, it needed to be high-fidelity preprocessed based on GAF technology. To achieve the integration of multi-source data from virtual and physical and multi-task prediction, the MTL architecture is constructed. The common features between multiple tasks can be identified and shared, and the comprehensive prediction of product shape can be achieved.

**4.2 Pre-processing of real-time measurement series**

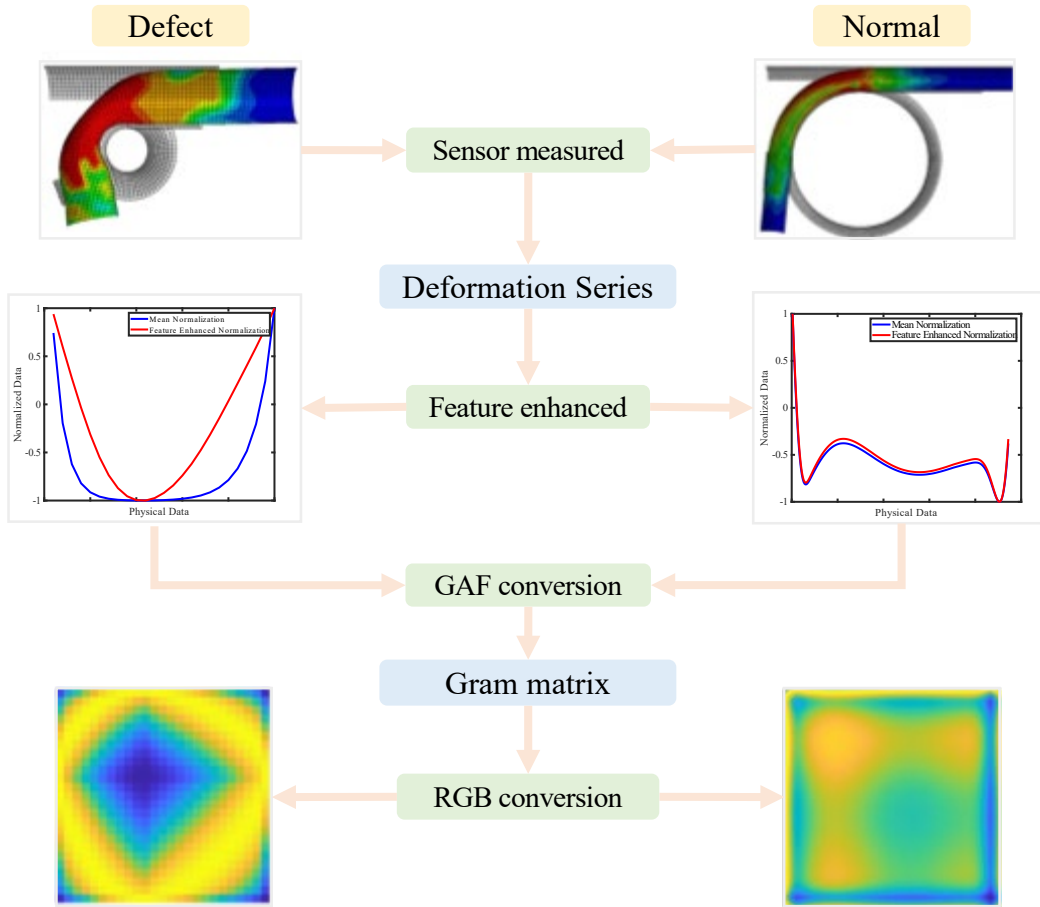

**Fig. 3** Pre-processing of physical series data with GAF conversion

The real-time deformation data transmitted by the sensor can reflect the influence of low-reliable acquire parameters and uncertain environment factors, which significantly reduces the dimension of input parameters and improves the flexibility and reliability of the prediction. However, the formats of the deformation series are not unified, making it difficult to be used in machine learning directly. Moreover, the common pre-processing methods, such as padding and series elongation, the real-time information is limited to a fixed period, which have the risk of data distortion and information loss. Recently, GAF has been proposed (Wang and Oates, 2015a) for the scaling and alignment of series information in image dimension, as shown in Fig. 3, which improves the reliability and fidelity of the physical data.

**4.2.1 GAF conversion**

The generation of tube bending defects is a process of gradual deformation, which can be reflected by the sensor measured data. GAF is a two-dimensional encoding method for the time series data. It can map the series to the image dimension, retains the complete series order and amplitude information, which improves the anti-interference ability for series scaling and alignment.

A rescaling of the real-time series observations needs to be conducted before the GAF conversion, which can be regarded as normalization to a specific interval, as shown in Eq. (1). In the original GAF method, this interval is [-1,1] by the mean normalization.

$$\widetilde{x^i} = \frac{(x_i - \max(X) + x_i - \min(X))}{\max(X) - \min(X)}, i = 1,2,\dots,n \tag{1}$$

However, the radial deformation is insignificant in the overall deformation of the tube bending. Besides, the potential wrinkling is located on the inner side of the bent-tube, bordering the bending die. It cannot be measured directly by the sensor. Eq. (1) is not conducive to reflecting the features of small fluctuations. Therefore, different from the original GAF, Eq. (2) is proposed to enhance the physical data features according to the characterization of the defect deformation.

$$\widetilde{x^i} = \frac{(e^{x_i} - e^{\max(X)}) + (e^{x_i} - e^{\min(X)})}{e^{\max(X)} - e^{\min(X)}}, i = 1,2,\dots,n \tag{2}$$

Where $X = \{x_1, x_2 \dots x_n\}$ is the physical deformation data.

The small deformation feature is enhanced by Eq. (2) while the original trend is ensured. As shown in Fig. 3, the small fluctuations of the original data are significantly enhanced, while the already existing obvious fluctuations are preserved. At the same time, the physical data can be normalized into [-1,1], which can be represented as $\widetilde{X} = \{\widetilde{x}_1, \widetilde{x}_2, \dots, \widetilde{x}_n\}$. As the original GAF method, the $\widetilde{X}$ then be converted to a polar coordinate system as shown in Eq (3).

$$\begin{cases} \emptyset_i = \arccos(\widetilde{x}_i), i = 1,2,\dots,n \\ r_i = \frac{t_i}{N}, t_i \in N \end{cases} \tag{3}$$

The Gram matrix and trigonometric sum with be implemented to realize the final GAF image conversion, as shown in Eq. (4).

$$G(x,y) = \begin{bmatrix} \cos(\emptyset_1 + \emptyset_1) & \cdots & \cos(\emptyset_1 + \emptyset_n) \\ \cos(\emptyset_2 + \emptyset_1) & \cdots & \cos(\emptyset_2 + \emptyset_n) \\ \vdots & \ddots & \vdots \\ \cos(\emptyset_n + \emptyset_1) & \cdots & \cos(\emptyset_n + \emptyset_n) \end{bmatrix} \tag{4}$$

The GAF can identify the temporal correlation between other points of the series through its own positional relationship, preserving the dependency between series. Furthermore, this bijective mapping can preserve the feature information of the series when the image is rescaled with a simple and efficient calculation.

### 4.2.2 Image format unification and RGB conversion

As the time-series data changes, the size of the GAF matrix will also change. Therefore, the transformed GAF image needs to be scaled to a uniform size before the data is used for machine learning. As shown in Eq. (5), $G$ can be scaled using bi-cubic interpolation (Keys, 1981) to achieve the best image scaling effect.

$$G_s(X, Y) = \sum_{i=0}^{3} \sum_{j=0}^{3} a_{ij} x^i y^j \tag{5}$$

where $a$ is the weighting coefficient, which depends on the characteristics of the interpolated data, $G_s$ is the scaled image.

Studies have shown that GAF images with color can perform better in CNN (Barra et al., 2020). In order to better preserve features, $G_s$ should be converted to RGB images before training (Wang and Oates, 2015a; Wang and Oates, 2015b). A *colormap* containing $m$ color combinations is determined at first. Specifically, the *colormap* is an $m \times 3$ matrix, and each row constitutes a set of RGB. $G_s^c$ is then obtained by rounding the result of $G_s$ linearly mapped to the range $[1, m]$. The color at $(X, Y)$ is finally determined by the RGB of the $G_s^c(X, Y)$th row of the colormap.

### 4.3 DT-enhanced Multi-source-input MTL

MTL has the ability to identify and share common features between different tasks. The springback and defects are essentially different behaviors of the same bending deformation, which are closely related tasks. Given the high cost of the defect dataset acquisition, enough MTBF information cannot always be guaranteed. Through MTL, the deformation knowledge and the underlying characteristics of each task can be shared to improve the training efficiency and prediction accuracy. At the same time, the DT data loop can provide physical and virtual data. The multi-source input architecture can integrate these different information sources, realizing the feedback of the real-time processing

### 4.3.1 Multi-source-input MTL architecture

The multi-source-input architecture is adopted to integrate physical data and virtual data. The proposed deep learning network mainly combined two independent networks with corresponding input layers for virtual data and physical data, respectively. Specifically, the defect is essentially the axial deformation, which can be directly reflected by physical data. As shown in Fig. 3, there is a significant difference between the normal and defects with GAF image. However, defect classification by virtual data requires dimension-based mapping. At the same time, benefiting from the apparent characteristics with a small mapping relationship (Zhou et al., 2021), the springback regression with the parameter-dimension is more efficient. In other words, the defect classification based on GAF image and the springback regression based on shape and design parameters is a better choice. It has also been verified in Section 5.3. On this basis, the soft parameter sharing mode is adopted, and the two tasks are relatively independent, while sharing common features.

Based on the determination of the overall logic of MTL, an appropriate feature extraction architecture needs to be adopted. Since the small mapping relationship between the virtual data and springback, the single-layer fully connected (FCN) layer can meet the requirement of feature extraction. The features of the images are more dependent on deep learning based on Convolutional Neural Network (CNN). The convolutional (Conv) layer can extract features of the input directly, and the batch normalization (BN) layer accelerates the training efficiency by reducing internal covariate shift. The rectified linear units (Relu) layer is taken as the activation function, and the max-pooling layer retains the main features while reducing the parameters. Specifically, *Conv1-BN1-MaxPooling1-Relu1* realizes the initial feature extraction, on this basis, *Conv2_1-Relu2_2* achieves the identification of a more complex feature. Particularly, there is only one max-pooling layer to preserve the features before sharing as much as possible. To improve the efficiency of backpropagation, *ConvSkip-BNSkip-ReluSkip* is constructed.

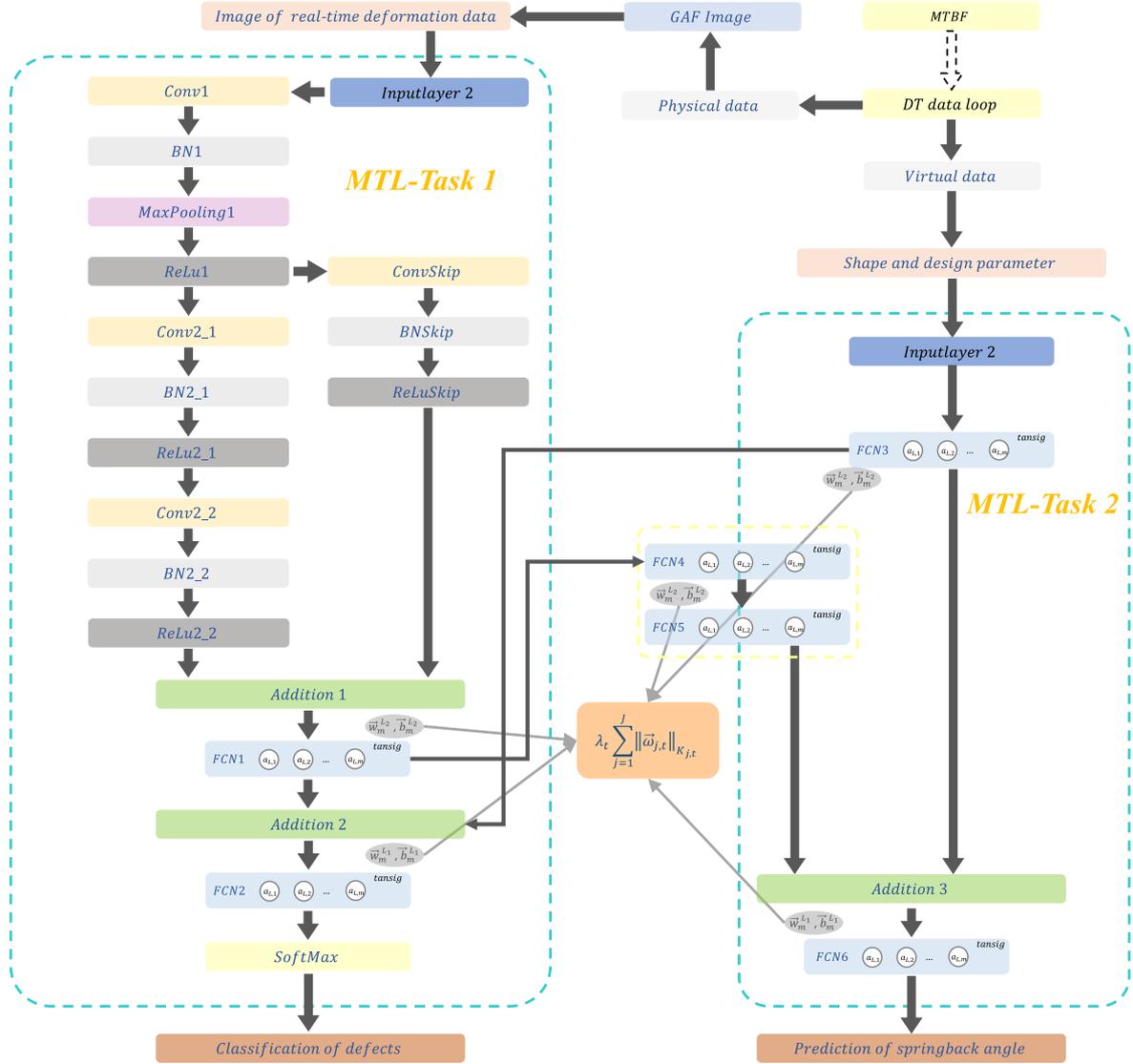

**Fig. 4** Multi-source-input MTL architecture

Shallow networks are versatile and reusable in feature extraction and can improve the performance of knowledge sharing (Neyshabur et al., 2020; Yosinski et al., 2014), which will be identified and shared. Specifically, as shown in Fig. 4, the general feature of virtual data extracted by *FCN3* is shared with MTL-Task1 through *Addition2*, determining the parameters of *FCN2* together with *FCN1*, and further determines the final classification of the defect. Similarly, the general feature of the preprocessed physical data extracted by *Conv1-Addtion1* is shared with MTL-Task2 through *FCN1-FCN4-FCN5*, determines the parameters of *FCN6* together with *FCN3*, and further determines the final springback prediction results. In this process, the group regularization with L1/L2 strategy is adopted to reduce the probability of model overfitting, improve the sparsity of feature sharing, and prevent information redundancy. Generally, the feature sharing layers are constrained by L2 regularization, and the feature accepting layers are constrained by L1 regularization.

The dataset will be randomly divided into three parts, i.e. validation set, testing set and training set. The hyperparameters of the network will be tuned with the validation set, while the test set verifies the final accuracy. The training set is reshuffled at each epoch, and is divided into multiple small batches, while each small batch has four sets of data. The Adam algorithm is adopted for parameter update, and the upper gradient limit is set to 1 to prevent the exploding gradients.

**4.3.2 Group regularization strategy on feature sharing and accepting layers**

As stated in Occam's razor, we should choose the one from all possible models that characterize the MTBF better and simpler. Regularization can prevent overfitting and improve model generalization performance by introducing additional information, keeping the sparse of the model. Compared with single-task learning, multi-task learning has more parameters, and there is a trend of cross-correlation between features. Due to the various information shared compositions, the feature connection between tasks should be sparsity-constrained. Considering the general regression problem with $J$ factors:

$$Y = \sum_{j=1}^{J} X_j \beta_j + \varepsilon, j = 1,2,\dots,J \tag{6}$$

Where Y is an $n \times 1$ vector, $\varepsilon \sim N_n(0, \sigma^2 I)$, $X_j$ is an $n \times p_j$ matrix corresponding to the $j$th factor and $\beta_j$ is the coefficient vector.

Based on that, the regularization method can be defined as

$$\hat{\beta}(\lambda) = \arg(\min_{\beta})(\|Y - X\beta\|^2 + \lambda \|\beta\|_K) \tag{7}$$

Where $\lambda$ is a hyperparameter and $\|\cdot\|_K$ represents the vector $K$-norm.

For $T$ tasks, the group regularization term of the $t$ th can be defined as:

$$\lambda_t \sum_{j=1}^{J} \|\beta_{j,t}\|_{K_{j,t}} \tag{8}$$

Where $\beta_{j,t} \in \mathbb{R}^{J \times T}$, represents the constraints on the $j$th row features of the $t$th task.

For both tasks, the feature sharing parameters will be L2 regularized, i.e. $K = l_1$, while the feature accepting parameters will be L1 regularized, i.e. $K = l_2$, as shown in Fig. 4. For the parameter sharing and accepting layer, $\lambda$=0.00025. Specifically, the L1 regularization (i.e., lasso regularization) constrains the $l_1$-norm of the network specific parameters to achieve the zeroing effect of the low-correlated items. This method improves the sparsity and ensures the simplicity of the model. The L1 regularization is applied on the feature accepting layers, i.e., the *FCN2* and *FCN6* in both tasks, reducing the information redundancy and ensuring the feature transmission efficiency. The L2 regularization (i.e., ridge regression) constrains the $l_2$-norm of the parameters. Different from the L1 regularization, L2 ensures sparsity by restricting the model space. The L2 regularization is applied on the feature sharing layers, i.e., the *FCN1* and *FCN3* in both tasks, improving the efficiency of information integration and ensuring optimal feature information sharing.

In addition, all other layer parameters are performed under L2 regularization with $\lambda$=0.0001 to ensure global sparsity and prevent overfitting. Combined with the batch normalization layer and max-pooling layer, the distance between model parameters can be regularized, reducing the overfitting probability and the

Rademacher complexity in this work.

## 5 Experiment and results

The 6061-Al alloy bent-tube has been widely used in industries such as aerospace and aviation due to its advantages of the lightweight, high strength, good corrosion resistance, and good processability. However, its springback is particularly significant, as well as other potential defect trends. It is necessary to apply the proposed DT-enhanced multi-source-input MTL method for accurate prediction to obtain the high-precision result.

To verify the effectiveness of the proposed method, the effects of equipment errors need to be excluded, such as out-of-sync communications caused by hardware latency (Tao et al., 2018a). The FE simulation has the ability to realize the complete construction of MTBF, and provide near-real processing and results of the product. Based on the verification with actual experimental, FE simulation-based scenery can be taken as the real physical processing scenery effectively, avoiding the influence of other irrelevant factors.

### 5.1 6061-Al tube bending experimental case study and data set construction

#### 5.1.1 6061-Al tube bending experiment

Before the MTBF is performed in FE simulation scenarios, 6061-Al is utilized to verify the accuracy of FE simulation. As shown in Fig. 5, the 6061-Al tube bending experiment was carried out using the KM-A50-CNC-320 CNC tube bending machine. The experimental settings are shown in Table 2. The MMX-2 friction and wear testing machine are used to measure the friction between the tube blank and the molds. After loading, the preload force on the tube blank caused by the clamp die and the pressure die is measured using the strain gauges, and the gap is measured with the three-coordinate interference detection machine. The FreescanX5 scanner is used to scan the final product. The scanned point cloud is imported into MATLAB for post-processing.

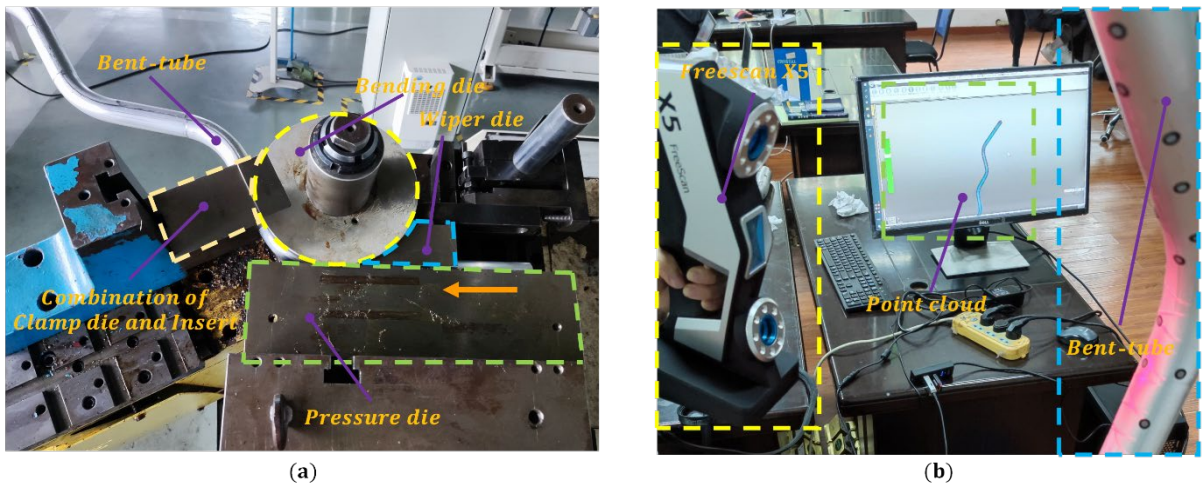

**Fig. 5** Metal tube bending experiment. (a) RDB processing. (b) bent-tube points cloud scanning

**Table 2** RDB experiment process plan

| case | Design parameter | Mold parameter |
| --- | --- | --- |

|   | $D$ /mm | $t$ /mm | $R_D$ /mm | $\alpha_D$ /rad | $\omega_B$ /rad·$s^{-1}$ | $L_P$ /mm | $v_B$ /rad·$s^{-1}$ | $f_B$ | $f_W$ | $f_P$ | $G_B$ /mm | $G_c$ /mm | $G_P$ /mm |
|---|---|---|---|---|---|---|---|---|---|---|---|---|---|
| 1 | 50 | 3 | 150 | 45 | | | | | | | | | |
| 2 | 50 | 3 | 150 | 50 | | | | | | | | | |
| 3 | 50 | 3 | 150 | 55 | 0.6 | 70 | 0.24 | 0.1 | 0.05 | 0.05 | 0.21 | 0.24 | 0.23 |
| 4 | 50 | 3 | 150 | 60 | | | | | | | | | |
| 5 | 32 | 3 | 150 | 60 | | | | | | | | | |
| 6 | 32 | 3 | 150 | 90 | | | | | | | | | |

### 5.1.2 FE simulation validation

The ABAQUS2016 is used as the FE simulation platform. The S4R shell elements with the 9th-order Simpson integration points in the thickness direction are used for tube modeling. The molds adopt R3D4 rigid units. Penalty contact is given between the tube blank and the molds. With velocity loading, the clamp die rotates with the bending die, and the tube blank is deformed under the constraint of friction and preload stress, as the same in actual processing. In the first stage, since the springback and the defects are related to the loading velocity, the explicit dynamic analysis is adopted in processing. In the second stage, after the removal of molds, since the springback is in a steady state, the implicit static analysis is adopted.

$$\begin{cases} AveL^i_{Experiment} = average(P^i_{location_E}) \ P^i_{location_E} \in \{Experiment\ point\ cloud\} \\ AveL^i_{FE\ simulation} = average(P^i_{location_F}) \ P^i_{location_F} \in \{FE\ simulation\ point\ cloud\} \end{cases} \quad (9)$$

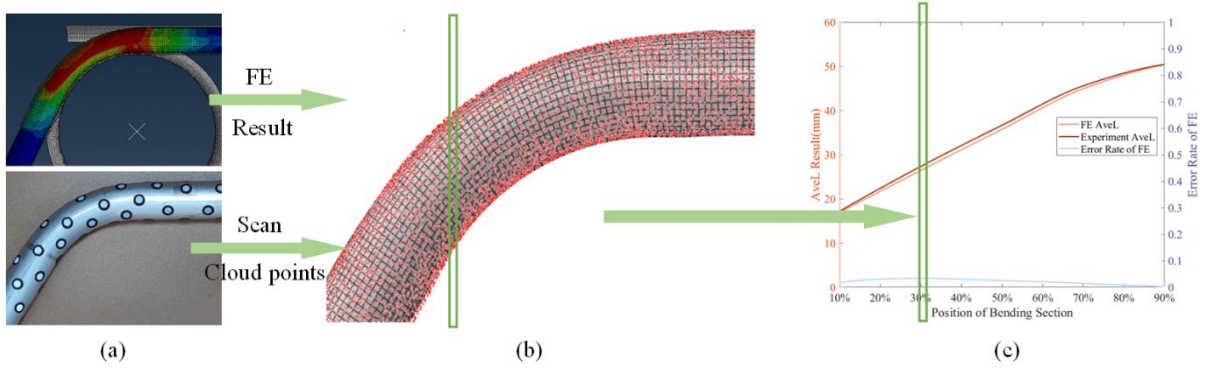

**Fig. 6** Final shape comparison of a set of processing plan. (a) Results of FE simulation and experiment. (b) the superimposition of the experimental point cloud and the FE simulation geometric results. (c) the $AveL$ deviation between the results of FE simulation and experiment.

According to the same settings of the experiment as in Table 2, FE simulation is carried out, and the result is compared with the tube bending experiment. From the perspective of the final shape, by proportionally superimposing the simulated geometric results with the scanned point cloud of the experiment bent-tube, intuitive product comparisons can be made, as shown in Fig. 6b. The average location of $ith$ green frame can be denoted by $AveL$, as Eq. (9) shows. The full bending section deviation is shown in Fig. 6c. From the perspective of the springback, compared with the experimental results, the average springback precision of the six FE simulation sets is 2.73%, and the average $AveL$ error is 3.15%.

The accuracy of the FE simulation in the final shape and forming index meets the requirements. It can be concluded that the FE simulation can realize the complete construction of MTBF, and provide near-real processing and results of the product.

### 5.1.3 Batch simulation data set construction

The proposed multi-source-input MTL should be able to integrate with any combination of shape parameter and the design parameter, and any physical data at any time. Therefore, the establishment of the data set should contain virtual data that can produce normal and defective samples and corresponding physical data.

Since the validation experiments are performed on physical experiment-verified FE simulation scenarios, the generation of the dataset is also based on batch simulation. The batch simulations are based on the uniformly distributed Latin hypercube sampling. The dataset contains parameters and real-time processing information to train the MTL using virtual data and physical data. Specifically, shape, design, and process parameters are obtained by sampling and submitted for simulation accordingly. Due to the randomness of the sampling parameters combination, the simulated result of a sample could be normal or defective. Its springback is also recorded. At the same time, the simulation processing is exported in real-time as the deformation data measured by the sensor, i.e., the physical data.

Physical data for a sample could be collected at any stage during processing. Specifically, for the $Mth$ sample, the $N$ frame states outputting are set uniformly. The $L_i\{i \in I\}th$ frame is randomly extracted as the sensor measures series of $Mth$ sample. The physical data reflect the deformation state at $(L/N) \times steptime,$ where $steptime$ is the total processing time. For different samples, physical data could be at different deformations state of processing. On this basis, the trained multi-source-input MTL has the ability to handle the physical data of any processing state, and achieve the real-time prediction of the MTBF.

The data set contains 1150 samples and is randomly divided into three sets. Among them, the validation set contains 150 samples, the testing set contains 150 samples, and the training set contains 850 samples. In each set, the "Normal" accounts for about 65%, the "Collapse" accounts for about 7%, the "Wrinkling" accounts for about 13%, and the "Collapse and Wrinkling" accounts for about 15%.

### 5.2 Multi-source-input MTL prediction results

The training result using the dataset in Section 5.1.3 is shown in Fig. 7. The bent-tube data set has more "Normal" and fewer "Defects", with a limited data set number, the classification accuracy can still reach 87.3%. It can be seen from the confusion matrix that in the case of misclassification, there is a possibility that the slight defect tendency under the label that should be normal is mislabeled as the defect attribute. It is because that the stress distribution cloud map generated by the Abaqus2016 discrete analysis can be misleading. At the same time, the deformation trend is gradual, so the possibility of one defect being misjudged as normal or another defect before it develops to explicit cannot be excluded. At the same time, these two types of defects may appear independently or simultaneously, and the proposed method can identify this difference.

The time dimension in the test group is analyzed as shown in Fig. 7b. It can be seen that with the increase of the time dimension, the prediction accuracy will be greatly increased. Specifically, the physical data in the data set may represent the deformation series at different processing stages. In the initial stage of processing, when the $(L/N)$ is small, the prediction error is relatively large, which may be because the deformation characteristics in the early stage of processing are not obvious and not reliable yet. When the processing approaches to completion, and the $(L/N)$ become larger, the reliability of the physical data increases and the prediction error decreases. In other words, for multiple samples, the closer the processing is to completion, the higher the confidence of physical data. In practical application, the real-time prediction and correction can be implemented based on the update of physical data. The proposed method has the ability to reach the accuracy of classification over 95% and the RMSE of regression below 0.6 with the physical data at the end of the processing. From another perspective, physical data has a great influence on the prediction of both defects and springback, also indicating that the multi-source-input MTL realizes the sharing of common features. Therefore, it can be concluded that with the increasing prediction accuracy, the multi-source-input MTL prediction method meets the requirements of industrial applications.

In practical applications, the machine tool control system can update the design parameter in real-time according to the prediction of the DT enhanced system. At the same time, the priority in processing is also placed on determining whether the potential defect occurs rather than determining the specific type of potential defect.

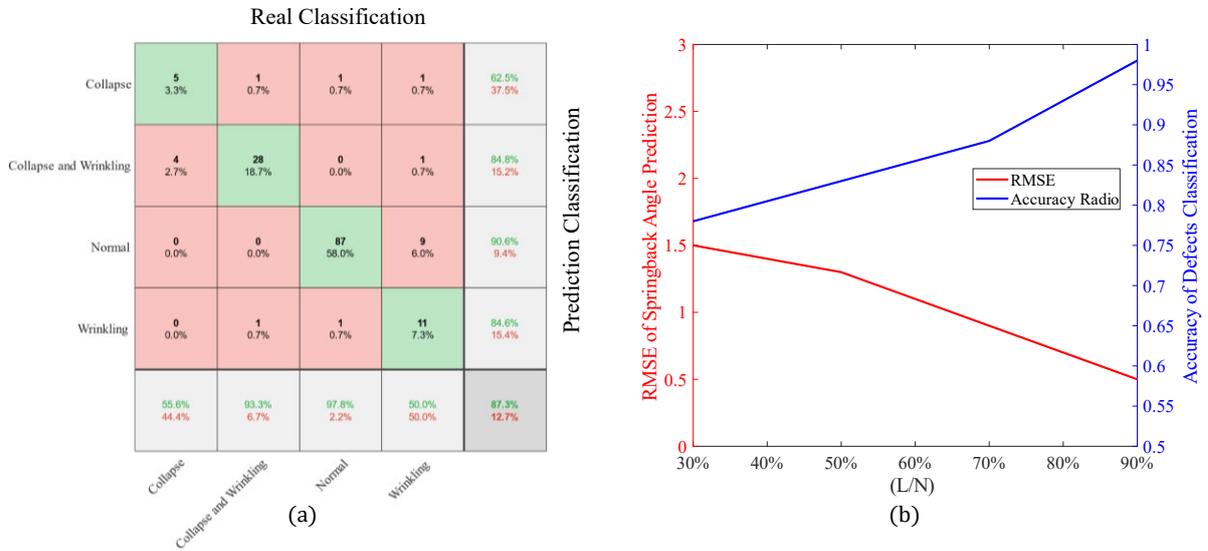

Fig. 7 Prediction accuracy of the Multi-source-input MTL. (a) The overall confusion matrix for the classification of defects. (b) Variation of the prediction results with the bending forming time

### 5.3 Comparison and discussion

GAF performs aligning transformation on series data in the image dimension, which can retain more features and improve prediction accuracy. To verify its superiority, the commonly used padding methods will be compared. It should be noted that the deep learning architecture of series prediction is different from that of image prediction. To control the variables, the padding method will be converted into images after aligning. The feature-enhanced GAF proposed in this work is marked as GAF-FE. The padding methods include padding at both ends, on the left and on the right, which are all commonly used methods, marked as padding-BE, padding-L, and padding-R, respectively. At the same time, the original GAF also be compared, terms GAF-original. The results are shown in Table 3.

Table 3 Effectiveness of GAF

|  | Best Accuracy | Ave. Accuracy | Best RMSE | Ave. Accuracy |
| --- | --- | --- | --- | --- |
| GAF-FE | 92.3% | 86.0% | 0.7699 | 1.1756 |
| GAF-original | 86.0% | 80.3% | 1.2756 | 1.5497 |
| padding-BE | 79.5% | 72.3% | 1.9775 | 2.4535 |
| padding-L | 72.0% | 65.4% | 2.6736 | 2.9113 |
| padding-R | 73.0% | 68.0% | 1.5317 | 2.5961 |

It can be seen that the padding methods of the two-end complement have higher accuracy in padding types, while the accuracy of GAF-FE with feature enhancement is higher than that of GAF-original.

The effectiveness of the L1/L2 group regularization strategy adopted in this work is also verified. L2 regularization only and L1 regularization only methods and no regularization are compared. The same dataset and training methods are used for the comparative experiments, and each is trained five times. The results are shown in Table 4. It illustrates that the accuracy of the L1/L2 combination is significantly higher than the other methods in terms of the best effect and network stability performance.

**Table 4** Effectiveness analysis of L1/L2 regularization

| | Best Accuracy | Ave. Accuracy | Best RMSE | Ave. Accuracy |
|---|---|---|---|---|
| L1/L2 Regularization | 92.3% | 86.0% | 0.7699 | 1.1756 |
| L1 Regularization | 85.0% | 82.3% | 1.2257 | 1.6345 |
| L2 Regularization | 87.5% | 83.3% | 1.2736 | 1.4172 |
| No-Regularization | 87.0% | 80.3% | 1.4799 | 1.5833 |

The multi-source-input MTL can efficiently utilize the common features between the two tasks and improve the training efficiency and accuracy. To verify its superiority, a comparison between multi-task and single-task learning is conducted. As mentioned in Section 4.3.1, the MTL-Task 1 structure is based on the CNN architecture, while the MTL-Task 2 structure is based on the FCN architecture. The two architectures are coupled with image and parameter input, respectively. Among them, the *SoftMax* layer will be removed when the MTL-Task 1 outputs the regression result, and the *SoftMax* layer will be added when the MTL-Task 2 outputs the classification result. The results are shown in Table 5.

**Table 5** Comparison of MTL and single-task learning

| | Multi-source-input MTL | MTL-Task 1 | | MTL-Task 2 | |
|---|---|---|---|---|---|
| Accuracy | 87.3% | 82.1% | - | 80.9% | - |
| RMSE | 0.7699 | - | 1.4409 | - | 1.7251 |
| Time | 3'03" | 1'49" | 1'27" | 0'35" | 0'32" |

It can be seen that the multi-source-input MTL has the highest regression and classification accuracy. The MTL-Task 1 achieves higher accuracy in the regression and guarantees a certain classification accuracy, but the time cost can be higher. The classification accuracy of the MTL-Task 2 is guaranteed, but its regression error is too high to use in engineering applications and real-time prediction cannot be achieved.

The common pre-trained networks are trained on more than a million images and are typically much faster and easier than training a network from scratch. Since the pre-trained image processing networks are specialized in classification problems, the comparison takes defects prediction as the main indicator. All pre-trained networks will use the same training parameters. The parameters of the first half of the pre-trained network layers are frozen to preserve the ability to extract image features. The parameters of the remaining layers are given a small learning rate to fine-tune the network parameters under the MTBF dataset. The output layer and its previous layer will be given a larger learning rate to improve the training efficiency. All training was performed on GTX1650s, and the results are shown in Table 6. Most of the state-of-art classification pre-trained networks are less accurate and higher time consuming for MTBF than that of the proposed multi-source-input MTL. Limited by the size of the dataset, it is difficult to train common pre-trained networks with high generalization and complex parameters, which limits their application in tube bending engineering.

Table 6 Results comparison of other pre-trained networks

|  | Proposed network | SqueezeNet | GoogleNet | Inception-v3 | ResNet-18 | ResNet-50 |
|---|---|---|---|---|---|---|
| Accuracy | 87.3% | 66.0% | 64.0% | 77.0% | 67.0% | 76.0% |
| Time | 3'03" | 2'19" | 4'30" | 16'18" | 2'27" | 8'31" |
|  | Res Net-101 | Inception-Res Net-v2 | AlexNet | MobileNet-v2 | ShuffleNet | DenseNet-201 |
| Accuracy | 77.0% | 74.0% | 71.0% | 76.0% | 70.0% | 80% |
| Time | 19'41" | 50'58" | 1'51' | 6'47" | 7'44" | 49'2" |

## 6 Conclusion

To comprehensively achieve the MTBF real-time prediction, a DT-enhanced prediction method based on multi-source-input MTL was proposed. The following remarkable conclusions are given as follows:

- Based on the MTL architecture, the error caused by complex deformation coupling in traditional mechanism analysis and machine learning methods was avoided. The MTL architecture identified and extracted common features in MTBF, and implemented comprehensive predictions for springback regression and defect classification at the same time.
- With technologies of the DT and GAF conversion, physical data was considered in MTBF prediction, and different lengths of series data were inputted into the machine learning architecture with high fidelity. The DT mapping connection between virtual and physical systems was achieved, the actual processing stage was reflected, and the real-time prediction was realized.
- In this process, the original GAF was improved according to the characteristics of the bent-tube deformation series data, and the feature enhancement of the small fluctuations was implemented. With the combined group regularization strategy, the accuracy of multi-source-input MTL and the efficiency of feature sharing were improved.

The effectiveness of the proposed method was verified on the physical experiment-verified FE simulation scenarios to exclude the effects of equipment errors. The accuracy of the simulation was first verified. Multiple control groups were compared to demonstrate the effectiveness of the improved GAF and the effectiveness of MTL architecture. At the same time, the common pre-training networks were compared with the proposed method. The results showed that the common pre-trained networks were less accurate and higher time consuming on MTBF prediction than that of the proposed multi-source-input MTL.

This work is expected to promote the development of high-precision prediction and processing of MTBF. However, some challenges still exist, such as the real-time process plan method for process plan adjustment. These challenges are expected to be solved by real-time optimization algorithms, which is also our further work.

**Funding**

This work is funded by the Joint Funds of the National Natural Science Foundation of China (U20A20287), the National Natural Science Foundation of China (51905476), and the Public Welfare Technology

**Competing Interests**

The authors declare that they have no conflict of interest.